\documentclass{article}
\usepackage{ijcai23}

\usepackage{times}
\usepackage{soul}
\usepackage{url}
\usepackage[hidelinks]{hyperref}
\usepackage[utf8]{inputenc}
\usepackage[small]{caption}
\usepackage{graphicx}
\usepackage{amsmath}
\usepackage{amsthm}
\usepackage{algorithm}
\usepackage[switch]{lineno}

\usepackage{wrapfig}
\usepackage{subfigure}
\usepackage{nicefrac}
\usepackage[algo2e,ruled,linesnumbered]{algorithm2e}
\usepackage{multirow}
\usepackage{booktabs,arydshln}
\usepackage{xcolor}

\title{Voting from Nearest Tasks: Meta-Vote Pruning of
Pre-trained Models for Downstream Tasks}

\author{
Haiyan Zhao$^1$
\and
Tianyi Zhou$^2$\and
Guodong Long$^1$\and
Jing Jiang$^1$\and
Chengqi Zhang$^1$
\affiliations
$^1$University of Technology Sydney\\
$^2$University of Maryland\\
\emails
Haiyan.Zhao-2@student.uts.edu.au,
zhou@umiacs.umd.edu,\\
\{guodong.long, jing.jiang, Chengqi.Zhang\}@uts.edu.au
}

\begin{document}

\maketitle

\begin{abstract}
As a few large-scale pre-trained models become the major choices of various applications, new challenges arise for model pruning, e.g., can we avoid pruning the same model from scratch for every downstream task? How to reuse the pruning results of previous tasks to accelerate the pruning for a new task? To address these challenges, we create a small model for a new task from the pruned models of similar tasks. We show that a few fine-tuning steps on this model suffice to produce a promising pruned-model for the new task. 
We study this ``meta-pruning'' from nearest tasks on two major classes of pre-trained models, convolutional neural network (CNN) and vision transformer (ViT), under a limited budget of pruning iterations.
Our study begins by investigating the overlap of pruned models for similar tasks and how the overlap changes over different layers and blocks.
Inspired by these discoveries, we develop a simple but effective ``Meta-Vote Pruning (MVP)'' method that significantly reduces the pruning iterations for a new task by initializing a sub-network from the pruned models of its nearest tasks.
In experiments, we demonstrate MVP's advantages in accuracy, efficiency, and generalization through extensive empirical studies and comparisons with popular pruning methods over several datasets.\looseness-1
\end{abstract}

\section{Introduction}
\label{sec:intro}

Large-scale pre-trained models usually contain tens of millions or even billions of parameters for promising generalization performance. 
The computation and memory of modern GPUs or clusters can support to train such models, but directly deploying them to edge devices can easily violate the hardware limits on memory and computation. Network pruning~\cite{han2015deep_compression,Tian2020MetaLearningWN,Li2020DHPDM,Chin2020TowardsEM} has been widely studied to compress neural nets by removing redundant connections and nodes. Numerous empirical results have verified that pruning can compress the original network into much smaller sub-networks that still enjoy the comparable performance.
Instead of reducing the network to the target size by one-time pruning, iterative pruning that alternates between pruning and fine-tuning for iterations usually achieves better performance~\cite{Han2015LearningBW,Li2017PruningFF}. 
Theoretically, a line of recent works~\cite{DBLP:journals/corr/abs-1803-03635,Ye2020GreedyOP,Savarese2020WinningTL,Malach2020ProvingTL} attempts to prove the lottery ticket hypothesis, i.e., the existence of such sub-networks, for different pruning settings.\looseness-1 

In a variety of practical applications, a large-scale pre-trained network like ResNet-50~\cite{He2016DeepRL} or Vision Transformer (ViT)~\cite{dosovitskiy2021an} usually needs to be pruned for a wide variety of devices and adapted to different downstream tasks. Running an iterative pruning algorithm for every device or task from the same pre-trained network can create enormous carbon footprint overload in our biosphere and waste a lot of computational power. 
On the other hand, the wide applications of a few pre-trained models have already created thousands of pruned models for different downstream tasks.
Can we reuse these pruned models as prior knowledge to save the pruning computation on new tasks?
We call this problem ``meta-pruning''. In this paper, we mainly focus on a special case of it, which initializes a sub-network for a given new task based on the pruned models of similar tasks.
Meta-pruning is non-parametric if no parametric model is trained to produce the initialization. It is analogous to MAML~\cite{Finn2017ModelAgnosticMF} in that the meta-objective optimizes the initialization of a network. It differs from MAML in that (1) both the sub-network's architecture and weights are initialized; and (2) the initialization is not universal but task-specific.\looseness-1 

Since meta-pruning aims to find better sub-network initialization for new tasks, we limit the iterations during meta-pruning to strengthen the impact of initialization on the final pruned model.
This also controls the computational cost and carbon footprint of meta-pruning much less than conventional pruning that requires many iterations.
Under this constraint, a well-performed pre-trained model is critical to the meta-pruning performance because (1) it needs to provide initialized sub-networks for different tasks; and (2) a few iterations of fine-tuning to the sub-networks should suffice to produce high-quality pruned models for targeted tasks.
Meta-pruning follows a practical setting where one single pre-trained model is tailored for different tasks using limited iterations. We study two classes of the most widely used pre-trained models, i.e., convolutional neural networks (CNN) and ViT.\looseness-1

The primary contribution of this paper is two folds.
In the first part, we conduct a thorough empirical study that applies  different pruning methods to CNN and ViT and compare their produced sub-networks for hundreds of downstream tasks.
No meta-pruning is studied in this part and its primary purpose is to (1) find the nearest tasks for a new task using different similarity metrics; and (2) compare the pruned models for different but similar tasks.
To this end, we build a dataset of tasks and their sub-networks pruned from the same pre-trained models.
Statistics and evaluations on this dataset indicate similar tasks with high similarity tend to share more nodes/filters/heads preserved in their pruned models, especially in deeper layers that notably capture high-level task-specific features.\looseness-1

Motivated by the empirical study, the second part of this paper proposes a simple yet strong meta-pruning method called ``meta-vote pruning (MVP)''. It can significantly reduce the pruning cost and memory required by previous pruning approaches yet still produce pruned models with promising performance.
Given a pre-trained model, MVP finds a sub-network for a new task by selecting nodes/filters/heads through majority voting among its nearest tasks, e.g., a filter will be sampled with a higher chance if it is selected into more sub-networks of similar tasks.
To keep the method simple, we sample the same proportion of nodes/filters/heads as the targeted pruning ratio. Then we apply a few iterations of fine-tuning to the initialized sub-network using training data of the new task.
Although a more sophisticated procedure can be developed, the proposed method already saves a substantial amount of computation and memory while maintaining a high test accuracy of pruned models. We demonstrate these via experiments over tasks from CIFAR-100~\cite{CIFAR}, ImageNet~\cite{5206848}, Caltech-256~\cite{griffin2007caltech} and several fine-grained datasets. The pruned models extracted from an ImageNet pre-trained model can also vote for tasks drawn from the unseen datasets with great performance, which shows the generalization of MVP.\looseness-1 

\section{Related works}

\paragraph{Network pruning}
Network pruning has been widely studied to compress network and accelerate its inference for a single task. We mainly summarize structure pruning below.
In CNN, to encourage the sparsity of the pruned network, $L_0$ \cite{Louizos2018LearningSN}, $L_1$ \cite{Liu2017LearningEC} or $L_2$ \cite{Han2015LearningBW} regularization have been used, and polarization regularization~\cite{Zhuang2020NeuronlevelSP} shrinks some nodes towards $0$ and meanwhile strengthen the others to keep important nodes intact.
Different criteria have been proposed to evaluate the importance of nodes/filters. \citeauthor{Li2017PruningFF} prunes filters with the smallest sum of parameters' absolute values. \citeauthor{Lin2018AcceleratingCN} prune filters according to the second-order Taylor expansion of the loss. Methods~\cite{bai2022dual,DBLP:journals/corr/abs-1803-03635} based on lottery ticket hypothesis try to find a well-performed sparse initialization for each task.\looseness-1

ViT has been widely used in computer vision and achieved SOTA performance in many tasks. The input patches for each block can be pruned to save computation for ViT. \citeauthor{goyal2020power} propose a metric for the importance of each patch and dynamically prune patches in each layer. PatchSlimming~\cite{tang2022patch} retains patches critical to preserve the original final output. HVT~\cite{pan2021scalable} is a CNN-like method which shortens the patch sequence by max-pooling. Another line of works~\cite{zhu2021vision,yu2022unified,yu2022width} automatically prunes the unimportant heads, nodes and blocks in ViT. These methods excel on single-task pruning but their cost linearly increases for multiple tasks (and thus more expensive than meta-pruning) because: (1) a large model needs to be trained for every task; (2) every task requires to prune its own large pre-trained model from scratch.
For both CNN and ViT, it is time-consuming for these pruning methods to build a pruned model for each unseen target task from a large pre-trained model.
Our proposed method can borrow the knowledge of the existing pruned models extracted by these pruning methods and use them to generate a well-performed pruned model for the unseen task with a few fine-tuning iterations.
\looseness-1

\paragraph{Meta-pruning}
To the best of our knowledge, \textbf{the non-parametric meta-pruning problem, i.e., how to prune a model for a target task using the pruned models of other tasks}, has not been specifically studied in previous work. However, several recent researches aim at learning meta(prior) knowledge that can improve pruning in other scenarios.
MetaPruning~\cite{Liu2019MetaPruningML} trains a weight-generation meta-network to prune the same network for the same task under different constraints, e.g., user/hardware defined pruning ratios. DHP~\cite{Li2020DHPDM} addresses the same problem but does not rely on any reinforcement learning or evolutionary algorithm since it makes the pruning procedure differentiable. Meta-learning has been studied to find better weight-initialization for pruning on different tasks, e.g., \citeauthor{Tian2020MetaLearningWN} applies Reptile~\cite{Nichol2018OnFM} for overfitting reduction. Meta-learning has also been studied to select the best pruning criterion for different tasks~\cite{He2019MetaFP}. In~\cite{Sun2020LearningSS}, a shared sparse backbone network is trained for multi-task learning but it cannot be adapted to new tasks.
Our method is the first one to use meta-learning to extract a pruned model for a new task. The main differences of our approach to them are: (1) we do not train a parametric meta-learner but instead use majority voting from similar tasks; and (2) our meta-voting generates a pruned small sub-network to initialize the target task training, which significantly reduces the pruning cost.\looseness-1

\section{Empirical Study: Pruning a Pre-trained model for Different Tasks}

In this section, we conduct an empirical study that applies  different methods to prune a CNN or ViT pre-trained model for over hundreds of tasks. Our study focuses on the overlap between the pruned models for different tasks and whether/how it relates to their similarity. To this end, we introduce different task similarities and compare the overlap associated with different similarity groups.
The results show that more similar tasks tend to share more nodes/filters/heads in their pruned models. And this holds across different pruning methods, datasets and pre-trained models.
No meta-pruning is used in the study.\looseness-1

\subsection{A Dataset of Pruned Models}
\label{sec:empirical_study}
While the number of possible downstream tasks and users can be huge in practice, the current progress on foundation models show that one or a few large-scale pre-trained models with light fine-tuning usually achieve the SOTA performance on most of them.
To simulate this scenario on a standard dataset, our empirical study creates a dataset of pruned models for hundreds of tasks from the same pre-trained model.
We choose CIFAR-100 and ImageNet for the study due to many classes in them.
For each dataset, we randomly draw $1000$ classification tasks, each defined on $5$ classes sampled without replacement.
We adopt ResNet-18~\cite{He2016DeepRL} pre-trained on CIFAR-100,  ResNet-50 and a small version of  DeiT~\cite{touvron2021training} pre-trained on ImageNet.
For ResNet-18 and ResNet-50, we prune two types of pre-trained models, i.e., the supervised training following~\cite{Devries2017ImprovedRO} and the self-supervised training following SimSiam~\cite{chen2020exploring} (only the encoder is used). For ViT, the training of its pre-trained model follows ~\cite{dosovitskiy2021an}.\looseness-1

\textbf{Iterative Pruning} We apply iterative filter-pruning (IFP) to ResNet. Unlike magnitude-based pruning~\cite{Li2017PruningFF} with one-time selection of nodes/weights, iterative pruning alternates between network pruning and fine-tuning of model weights for multiple iterations, each of which prunes $p\%$ of the remaining nodes/weights so it progressively prunes a large network to the targeted size. It usually performs better than other pruning methods and has also been mainly studied in theoretical works about Lottery Ticket Hypothesis~\cite{DBLP:journals/corr/abs-1803-03635}.
We take the activation values of filters averaged over all training samples to measure the importance of filters~\cite{molchanov2016pruning}, referred as Activation Pruning, in which filters with smaller activation values contain less information of input data.
The detailed procedure of IFP is described in Appendix.\looseness-1

\textbf{Automatic Pruning} Inspired by the SOTA ViT structured pruning method \cite{yu2022unified}, we prune ViT by automatic head\&node pruning (AHNP) for a given task, which parameterizes the sub-network as the pre-trained model with a learnable score multiplied to each prunable head and node. To encourage sparsity, the differentiable scores of all prunable heads and nodes are optimized with an additional $L1$ regularization loss. After each optimization step, we apply a simple thresholding to these scores to remove heads and nodes with small scores. The optimization stops if the pruned model reaches the targeted size and the model will be fine-tuned for a few iterations.
The detailed procedure of AHNP can be found in Appendix.\looseness-1

For tasks of CIFAR-100, we run IFP for all $1000$ tasks on ResNet-18.
And we apply IFP and AHNP to tasks of ImageNet on ResNet-50 and ViT respectively.
Finally, we create a dataset of pruned models for thousands of tasks over different pre-trained models.
For each task $i$, we record its labels $C_i$, the set of preserved nodes/filters/heads $\{\Omega_\ell\}_{\ell=1:L-1}$ and the pruned model $\theta_T$. We use the same hyper-parameters for different tasks.
For IFP on ResNet, we use a learning rate of $0.005$, pruning iterations of $1000$ and batch-size of $128$ for both the tasks of CIFAR-100 and ImageNet.
When applying AHNP to ViT, we follow the ViT training in \cite{pmlr-v139-touvron21a}. We reduce the pruning iterations to $1000$ and use a small learning rate of $0.00005$ for parameters inherited from the pre-trained ViT (to preserve its knowledge) and a large learning rate of $0.05$ for the learnable scores. The pruning ratio is $90\%$ for all pruned models. Refer to the Appendix for the detailed discussion about the computational cost of the model zoo.\looseness-1

\begin{figure*}[htp]
\vspace{-1em}
     \centering
         \includegraphics[width=\textwidth]{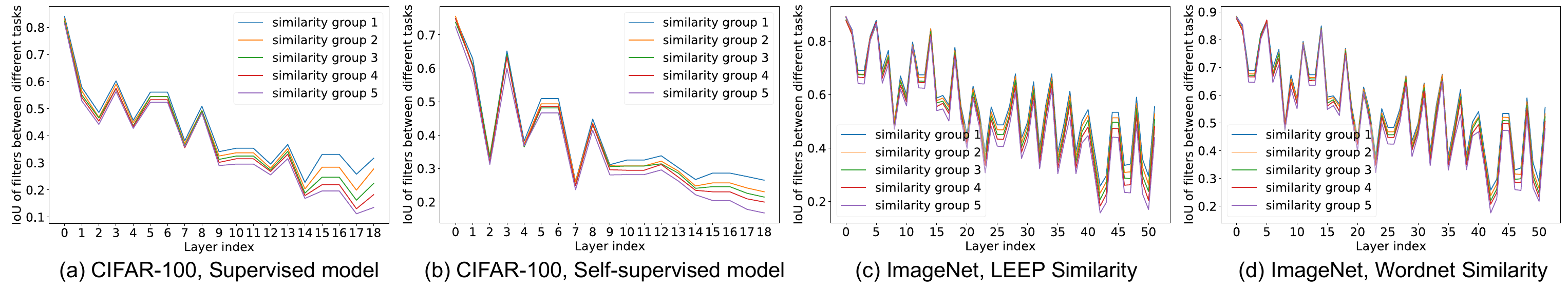}
\vspace{-2em}
\caption{IoU of layers in ResNet between tasks with different similarities.}
\label{fig:CNN_iou}
\vspace{-.5em}
\end{figure*}

\begin{figure*}[htp]
\vspace{-1em}
     \centering
     \subfigure[ImageNet, LEEP Similarity]{
         \centering
         \includegraphics[width=0.43\textwidth]{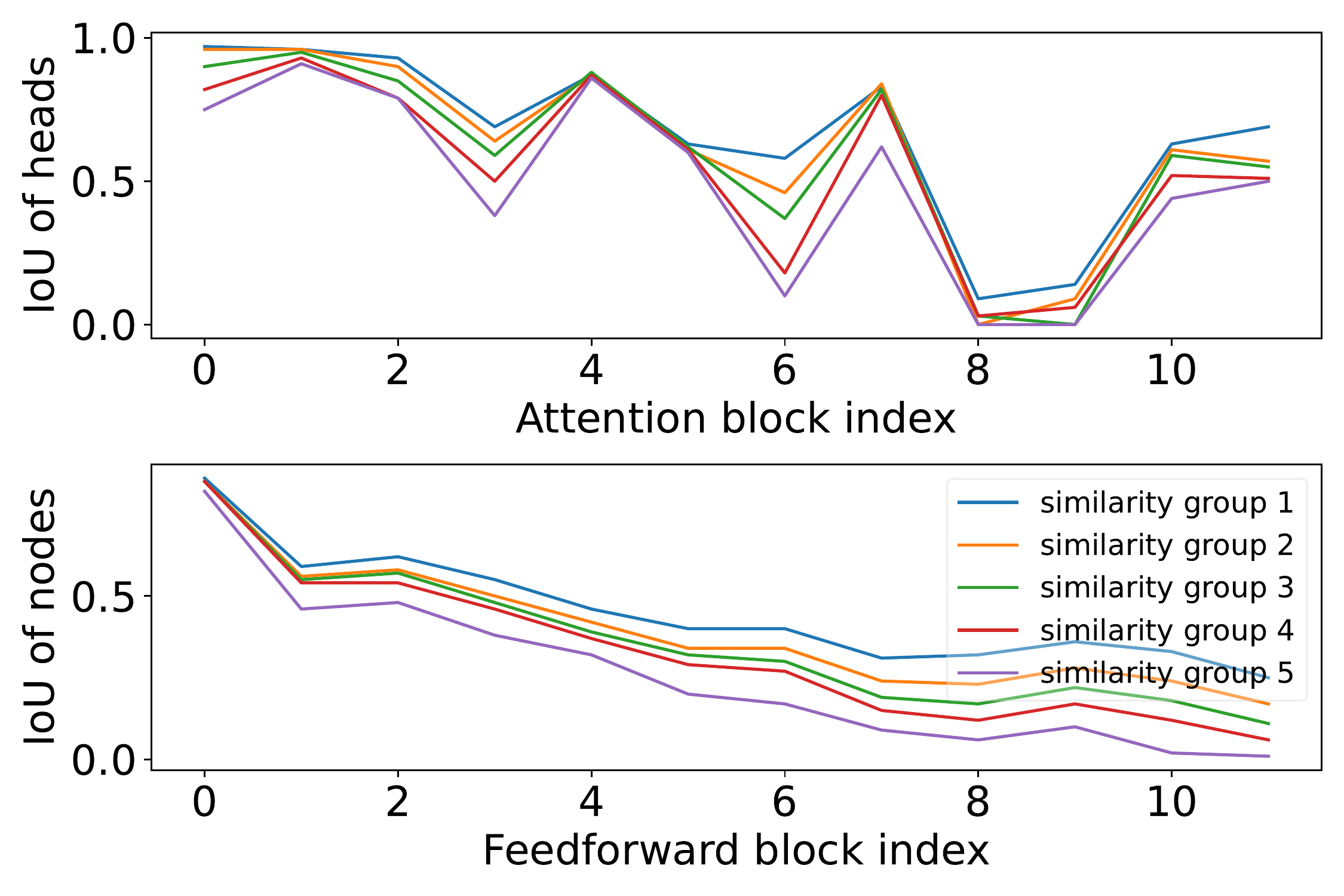}
     }
     \subfigure[ImageNet, Wordnet Similarity]{
         \centering
         \includegraphics[width=0.43\textwidth]{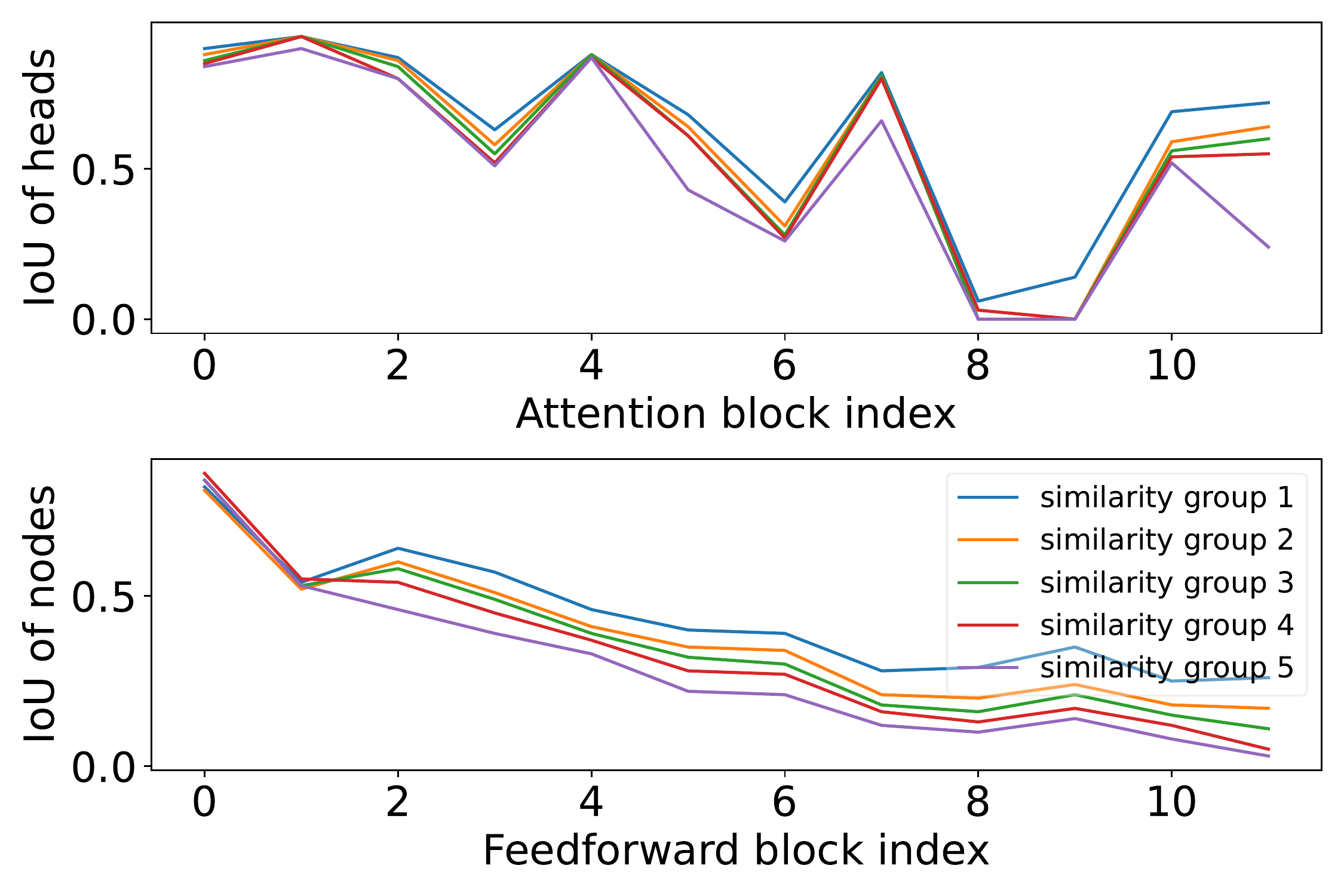}
     }
     \vspace{-1em}
\caption{IoU of blocks in ViT between tasks with different similarities.}
\label{fig:vit_iou}
\vspace{-1.5em}
\end{figure*}

\subsection{Do similar tasks share more nodes/filters/heads on each layer of their pruned models?}
\label{sec:task_similarity}
The representations learned for a task can still be helpful to its similar tasks. This motivates transfer/multi-task/meta learning methods. But do similar tasks also share more structures in their pruned sub-networks? 
We apply two metrics to measure the similarity between  classification tasks in our dataset and study whether/how the similarity relate to their shared nodes/filters/heads in different layers of their pruned models.\looseness-1

\textbf{Similarity Metrics} 
We apply two metrics to compute the similarity between tasks and find the nearest tasks, i.e., the Log Expected Empirical Prediction (LEEP)~\cite{nguyen2020leep} and the Wordnet wup similarity~\cite{pedersen2004Wordnet,wu1994verb}. LEEP score is widely used in transfer learning to estimate the knowledge transferability from a source task to a target task. In our study, for each target task, we can rank the other tasks by their LEEP similarity score from each of them to the target one. 
Computing the LEEP score only requires a single forward pass of the pruned model on the target task's data.
Wordnet wup similarity only requires the semantic labels of classes in each task and it is based on the depths of the their corresponding synsets in the Wordnet~\cite{miller1995wordnet} taxonomies. It does not depend on the pruned model so it is more efficient to compute.\looseness-1

\textbf{Overlap Between Tasks} 
Let $\Omega^i_\ell$ and $\Omega^j_\ell$ denote the sets of filters/nodes/heads remained in layer-$\ell$ after running IFP or AHNP for task $i$ and $j$ (when using the same pre-trained model), we measure the overlap of the two sets by intersection over union (IoU) ratio~\cite{EtudeDe}, i.e., ${\rm IoU}=\nicefrac{|\Omega^i_\ell\cap \Omega^j_\ell|}{|\Omega^i_\ell\cup \Omega^j_\ell|}$.\looseness-1

Fig.~\ref{fig:CNN_iou} (ResNet) and Fig.~\ref{fig:vit_iou} (ViT) report the IoU of each layer/block for pairs of tasks with different similarities.
For each target task, the tasks in the dataset are partitioned into $5$ similarity groups according to their LEEP scores or Wordnet similarities to the target task. The similarity decreases from group 1 to group 5. Specifically, for each test task, we compute its similarity scores with its neighbours in the model zoo. We sort these similarity scores and partition them into five groups of equal intervals. Neighbours whose similarity scores fall into a certain interval will be assigned to the corresponding group.\looseness-1

For all the datasets and architectures, \textbf{more similar tasks tend to share more filters/nodes/heads} (larger IoU) between their pruned models. 
Therefore, for a new task, the pruned models of its nearest  tasks preserve many important filters for it and combining them might result in a better and much smaller sub-network to  initialize the new task.
Moreover, for deeper layers/blocks in both ResNet and ViT, the gap between different similarity groups on the IoU increases because the features are more task-specific in deeper layers. 
Due to the same reason, for every similarity group, IoU decreases with depth in the overall trend (though fluctuating locally). 
Furthermore, Fig.~\ref{fig:vit_iou} shows that the IoU gap between similarity groups defined by the LEEP score is larger than that obtained by Wordnet similarity (refer to Appendix for detailed results). This indicates that the semantic similarity between class labels might not be as accurate as the LEEP score that takes the pruned model and its learned representations into account.\looseness-1

\begin{algorithm}[H]
    \setcounter{AlgoLine}{0}
    \SetKwInOut{Input}{Input}
    \SetKwInOut{Output}{Output}
    \SetKwInOut{Init}{Initialize}
    \Input{Target task $i$ and its training set $D_i$, pruning ratio $r$, $J$, $N$, a dataset of pruned models for different tasks}
    \Output{A pruned model for target task-$i$}
    \Init{$\Omega_\ell\leftarrow\emptyset$, the set of filters in layer-$\ell$}
    Sample/find $N$ similar tasks $N^i$ to task $i$ according to LEEP score or Wordnet similarity;\par
    \For{$\ell \gets 1$ \textbf{to} $L-1$}{
        Sample $(1-r)n_\ell$ filters with probability $p(k)$ (Eq.~(\ref{equ:vote_prob})) and add them to $\Omega_\ell$;\par
        \For{$k \in \Omega_\ell$}{
            Initialize filter-$k$ by averaging its parameters of tasks in $\{j\in N^i:k\in\Omega^j_\ell\}$;\par
           }
    }
    Fine-tune the pruned model for $J$ iterations on $D_i$.\par
\caption{\sc{Meta-Vote Pruning (MVP)}}
\label{alg:meta-pruning}
\end{algorithm}

\section{Meta-Vote Pruning (MVP)}

\begin{figure}
\vspace{-1em}
\centering
\includegraphics[width=0.5\textwidth]{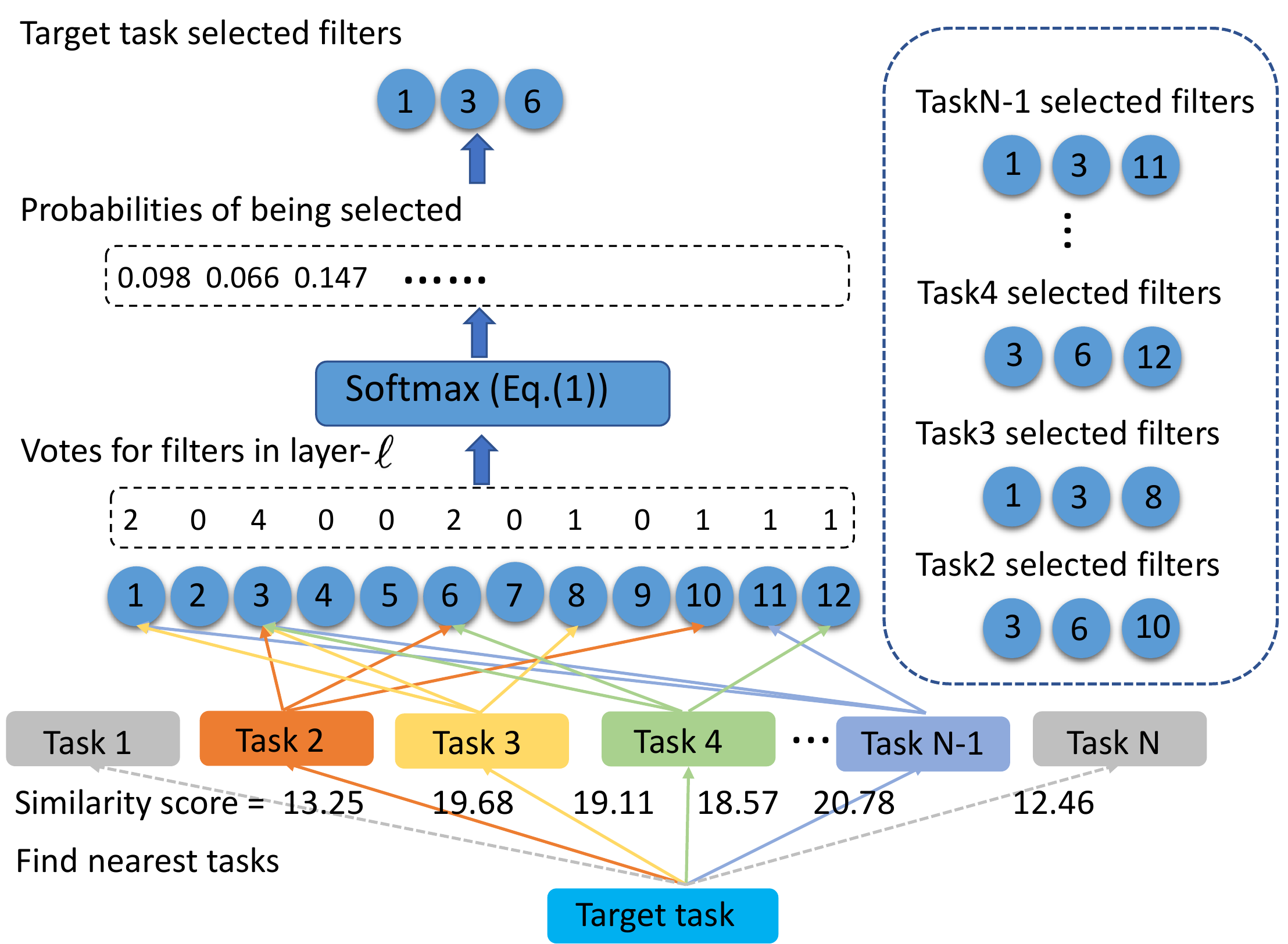}
\vspace{-2em}
\caption{The example of majority voting in MVP. Each similar neighbour task of the target task vote for filters which are reserved by its pruned model. Then, softmax is applied to the votes of all filters in layer-$\ell$ and filters with more votes have higher probability to be selected by the target task.\looseness-1 }
\label{fig:voting_flow}
\end{figure}

Inspired by the empirical study above, we propose a simple yet strong baseline ``meta-vote pruning (MVP)'' (Alg.~\ref{alg:meta-pruning}) for non-parametric meta-pruning. The procedure of MVP majority voting is shown in Fig.~\ref{fig:voting_flow}.
Given a target task $i$, MVP draws a sub-network of a pre-trained network by sampling filters/nodes/heads in each layer using majority voting from its nearest tasks $N^i$ and their pruned models. In particular, for each filter-$k\in[n_\ell]$ from layer-$\ell$ of the pre-trained model,
we apply softmax (with temperature $\tau$) to the
times of each filter being selected by tasks in $N^i$, which yields a probability distribution over all the filters $[n_\ell]$, i.e., $\forall k\in [n_\ell]$,\looseness-1
\begin{equation}\label{equ:vote_prob}
p(k)=\frac{\exp(|\{j\in N^i:k\in\Omega^j_\ell\}|/\tau)}{\sum_{h\in[n_\ell]}\exp(|\{j\in N^i:h\in\Omega^j_\ell\}|/\tau)}
\end{equation}
To initialize layer-$\ell$ of the sub-network, MVP samples filters from this distribution (without replacement) according to the targeted pruning ratio $r$. 
We further initialize the parameters of each filter-$k$ by averaging its parameters in the pruned models of the similar tasks which preserve filter-$k$. MVP then fine-tunes the initialized sub-network for a few iterations on the training set of the target task since MVP targets to keep the computational cost low.\looseness-1

\section{Experiments}

In this section, we conduct extensive experiments on CIFAR-100 and ImageNet over different pre-trained models, which evaluate MVP (Alg.~\ref{alg:meta-pruning}) and compare it with SOTA pruning methods under different settings.
We validate the strong generalization of MVP by applying it to unseen tasks from Caltech-256 and other fine-grained datasets.
We further study the effect of different pruning iterations, neighbour numbers, task sizes and similarity metrics for MVP. All the results show that MVP can outperform other methods with better performance and higher efficiency.
\looseness-1

\subsection{Implementation Details}

The experiments of MVP are mainly based on the tasks from the dataset introduced in Sec.~\ref{sec:empirical_study}.
For each setting of experiments, we randomly draw $100$ test tasks (i.e., the target task in Alg.~\ref{alg:meta-pruning}) from the dataset and treat the rest tasks as training tasks.
To evaluate MVP on CNN, we run MVP on the pruned models of ResNet-18 and ResNet-50 for CIFAR-100 and ImageNet respectively. For both these two experiments, we use the meta-pruning iterations of $100$, batch size of $128$, learning rate of $0.01$ and optimizer of SGD with the cosine-annealing learning rate schedule.
For experiments of ViT, MVP is applied to the pruned models of ViT for ImageNet. The meta-pruning iterations and batch size are also set as $100$ and $128$ respectively. Following the setting of training ViT in \cite{pmlr-v139-touvron21a}, we apply a small learning rate of $0.0002$ and optimizer of AdamW with the cosine-annealing learning rate schedule.
The small number of meta-pruning iterations demonstrates the efficiency of MVP.
The target pruning ratio of MVP for all tasks is $90\%$. All the results of accuracy shown in this section are averaged over the $100$ test tasks.\looseness-1

\subsection{Baseline Methods}
We compare MVP with several baselines and SOTA pruning methods.
We first implement two baselines to show the advantages of MVP.
(1) Conventional pruning. We apply a larger number of pruning iterations to extract pruned models for each target task by IFP or AHNP introduced in Sec.~\ref{sec:empirical_study}.
This baseline can be regarded as the upper bound performance.
(2) Random pruning. To validate whether the initialization of MVP makes sense, for each target task, we initialize its sub-network by randomly sampling the same number of nodes/filters/heads as MVP from the pre-trained model. We take this baseline as the lower bound performance.\looseness-1

We also include other SOTA pruning methods. 
For MVP on CNN, we compare MVP with IHT-based Reptile~\cite{Tian2020MetaLearningWN}, a meta-pruning method that uses Reptile~\cite{Nichol2018OnFM} and iterative pruning to find better weight-initialization for a pruned meta-model. Given a new task, it fine-tunes the pruned meta-model for a limited number of iterations to obtain the final pruned model.
MEST~\cite{yuan2021mest} is the SOTA method in sparse training community, which trains a model from a sparse sub-network so that less computation is required. 
DLTH~\cite{bai2022dual} is based on a variant of the Lottery Ticket Hypothesis. It transforms random tickets into winning tickets.
We compare MVP with UVC~\cite{yu2022unified} and PoWER~\cite{goyal2020power} on ViT pruning. Unlike AHNP, which prunes heads and nodes, UVC also skips the unimportant layers and blocks in ViT. Unlike parameter pruning, PoWER adopts a dynamic method pruning the input patches of each block for each input sample.
For a fair comparison, except for the upper bound baseline, the pruning iterations of all other methods and MVP are set to $100$. And the pruning ratios of all methods are set to $90\%$.\looseness-1

\begin{table}[htbp]
\vspace{-0.5em}
\begin{center}
\caption{\footnotesize Comparison between MVP and baseline methods on CNN. The '-SSL' behind each method means applying this method to pruned models extracted from self-supervised pre-trained models.  
\textbf{Bold} and \textcolor{gray}{\underline{\textbf{Bold gray}}} mark the best and second best accuracy.
}
\vspace{-0.5em}
\resizebox{\columnwidth}{!}{
	\begin{tabular}{lccccc}
		\toprule
		\multirow{2}*{\textbf{Methods}} & \multirow{2}*{\textbf{Pruning Iterations}} & \multicolumn{2}{c}{\textbf{ResNet-18}} & \multicolumn{2}{c}{\textbf{ResNet-50}} \\
		 &  & Acc & FLOPs & Acc & FLOPs  \\
		\midrule
		IFP & 1000 & \textcolor{gray}{\underline{\textbf{87.99$\pm$0.47}}} & 14.88(T) & \textcolor{gray}{\underline{\textbf{91.16$\pm$0.68}}} & 110.06(T)\\
		IFP-SSL & 1000 & 85.22$\pm$0.52 & 14.88(T) & 85.84$\pm$0.75 & 110.06(T)\\
		Random Pruning & 100 & 33.12$\pm$6.47 & 0.43(T) & 22.42$\pm$3.92 & 3.16(T)\\
		IHT-based Reptile\cite{Tian2020MetaLearningWN} & 100 & 75.23$\pm$0.87 & 0.43(T) & 73.40$\pm$0.75  & 3.16(T)\\
		MEST\cite{yuan2021mest} & 100 & 76.28$\pm$0.82  & 0.47(T) & 66.25$\pm$2.33 & 3.48(T)\\
		DLTH\cite{bai2022dual} & 100 & 74.46$\pm$1.24   & 4.28(T) & 69.33$\pm$1.56 & 31.64(T)\\
		MVP(\textbf{ours})  & 100 & \textbf{88.98$\pm$0.38} & 0.43(T) & \textbf{91.80$\pm$0.26} & 3.16(T)\\
		MVP-SSL(\textbf{ours})  & 100 & 86.82$\pm$0.13 & 0.43(T) & 85.92$\pm$0.26 & 3.16(T)\\
		\bottomrule
	\end{tabular} }
	\label{tab:cnn_results}
\vspace{-1em}
\end{center}
\end{table}

\begin{table}[htbp]
\vspace{-0.5em}
\begin{center}
\caption{\footnotesize Comparison between MVP and baseline methods on ViT. \textbf{Bold} and \textcolor{gray}{\underline{\textbf{Bold gray}}} mark the best and second best accuracy}
\vspace{-0.5em}
\resizebox{0.5\textwidth}{!}{
	\begin{tabular}{lccc}
		\toprule
		\multirow{2}*{\textbf{ Methods}} & \multirow{2}*{\textbf{Pruning Iterations}} & \multicolumn{2}{c}{\textbf{ViT}}  \\
		 &  & Acc & FLOPs  \\
		\midrule
		AHNP & 1000 & \textbf{89.48$\pm$0.62} & 81.50(T) \\
		Random Pruning & 100 & 58.71$\pm$4.14 & 3.25(T)\\
		UVC\cite{yu2022unified} & 100 & 80.30$\pm$0.57 & 26.73(T)\\
		PoWER\cite{goyal2020power} & 100 & 77.76$\pm$1.18 & 20.86(T)\\
		MVP(\textbf{ours})  & 100 & \textcolor{gray}{\underline{\textbf{89.23$\pm$0.49}}} & 3.25(T) \\
		\bottomrule
	\end{tabular} }
	\label{tab:vit_results}
\vspace{-1em}
\end{center}
\end{table}

\subsection{Main Results}
The results of applying MVP to tasks from CIFAR-100(ImageNet) on ResNet-18(ResNet-50) supervised and self-supervised pre-trained model, and the baseline methods are reported in Tab.~\ref{tab:cnn_results}.
On both datasets and pre-trained models, MVP outperforms IFP which spends 10$\times$ iterations of MVP. Hence, MVP can produce a higher-quality pruned model when using fewer iterations.
The results demonstrate that MVP can work well on tasks from both supervised and self-supervised pre-trained models.
The random pruning performs much poorer than MVP, which indicates the importance of majority voting from nearest tasks in selecting filters.
\looseness-1

We also compare MVP with SOTA pruning methods for CNN.
IHT-based Reptile~\cite{Tian2020MetaLearningWN} trains a universal sparse sub-network for all target tasks by applying meta-learning on training tasks.
MVP achieves higher accuracy than IHT-based Reptile under the same training iterations, implying that MVP can find an accurate sub-network for each target task as its initialization and improve its performance. 
MEST~\cite{yuan2021mest} can speed up pruning by starting training from a well-designed sub-network. As a variant of Lottery Ticket Hypothesis, DLTH~\cite{bai2022dual} proposes a method to transform any random ticket into the winning ticket.
MVP outperforms MEST and DLTH by a large margin because MVP is trained on a sub-network selected using meta knowledge from similar tasks. In contrast, the initial sub-network for MEST or the winning ticket of DLTH does not leverage any prior knowledge about the target task.\looseness-1

Tab.~\ref{tab:vit_results} shows the comparison between MVP and baseline methods on ViT.
Similar to the results on pruning CNN, the performance of MVP on ViT is comparable to AHNP that applies much more pruning iterations. The accuracy of random pruning is still much worse. 
MVP also outperforms SOTA pruning methods developed for ViT. 
Hence, on ViT, MVP can efficiently produce a small yet high-quality sub-network for each new task by exploiting the nearest tasks' models.
The baselines are slower and require more iterations than MVP because they need to re-train the model to achieve a small loss when some parameters or patches are removed.
Both UVC and PoWER cannot recover the accuracy under this strong constraint. 
In contrast, the majority voting in MVP directly produces a small sub-network from similar tasks' models so only a few iterations suffice to reach a downstream task performance comparable to AHNP with 10x iterations.\looseness-1

\begin{table}[ht]
\vspace{-0.5em}
\begin{center}
\setlength{\tabcolsep}{2.pt}
    \caption{Accuracy of MVP on unseen tasks.}
    \vspace{-0.5em}
\resizebox{\columnwidth}{!}{
	\begin{tabular}{lcccccc}
		\toprule
		\multirow{2}*{\textbf{Methods}}  & \multicolumn{3}{c}{\textbf{Caltech-256}}  & \multicolumn{3}{c}{\textbf{CUB200-2011}}  \\
		 & Iters & Acc & FLOPs & Iters & Acc & FLOPs  \\
		\midrule
		IFP & 800 & 80.28$\pm$1.64 & 93.06(T) & 800 & 77.09$\pm$0.51 & 23.99(T) \\
		IFP & 60 & 42.90$\pm$3.79 & 6.73(T)     & 80 & 52.85$\pm$2.95 & 2.17(T)\\
		MVP(\textbf{ours}) & 60 & \textbf{80.72$\pm$0.64} & 1.90(T)     & 80 & \textbf{79.54$\pm$0.88} & 0.63(T)\\
		\bottomrule
	\end{tabular} }
 \vspace{-1em}
\label{tab:caltech_results}

\end{center}
\end{table}

\subsection{Performance on Unseen Dataset}

In this section, to validate the generalization of MVP, we apply MVP to produce pruned models for target tasks from unseen dataset Caltech-256~\cite{griffin2007caltech} and fine-grained dataset CUB200-2011~\cite{wah2011caltech}, Oxford Flowers-102~\cite{nilsback2008automated} and Oxford-IIIT Pets~\cite{parkhi2012cats}, using the pruned models of tasks from ResNet-50 training on ImageNet.
The data of these datasets are never seen by the pre-trained model and tasks in the pruned model dataset. Each target task is defined on $5$ classes sampled without replacement from each dataset. \looseness-1

The performance of MVP on Caltech-256 is shown in Tab.~\ref{tab:caltech_results}, which is still comparable to the IFP using $10$x pruning iterations. When the number of pruning iterations of IFP decreases, its performance becomes much worse.
Besides Caltech-256, we also validate the effectiveness of MVP on more difficult fine-grained datasets, i.e., CUB200-2011, Oxford Flowers-102 and Oxford-IIIT Pets where images in different classes are from various species of birds, flowers and animals, which are hard to distinguish. On target tasks from fine-grained datasets, MVP also works better than IFP which needs much more computation costs. Results of Oxford Flowers-102 and Oxford-IIIT Pets can be found in Appendix.\looseness-1

The results show that MVP can still produce a high-quality initialization for the task from unseen datasets by majority voting of similar tasks so that the pruned model can converge quickly with high accuracy. MVP's great performance on fine-grained datasets implies that MVP can learn from different objects to facilitate the classification of hard-to-distinguish target tasks.
This experiment demonstrates that MVP can be applied to various datasets and generalizes well.
\looseness-1

\begin{table}[ht]
\begin{center}
\caption{\footnotesize Results on sub-tasks of different sizes for CIFAR-100.}
\resizebox{\columnwidth}{!}{
	\begin{tabular}{lcccccc}
		\toprule
		\multirow{2}*{\textbf{Methods}} & \multicolumn{3}{c}{\textbf{$10$-classification}} &
		\multicolumn{3}{c}{\textbf{$3$-classification}}   \\
		 & Iters & Acc & FLOPs &Iters& Acc & FLOPs  \\
		\midrule
		IFP & 1500 & 84.29$\pm$0.26 & 25.48(T) & 500 & 88.75$\pm$0.71 & 5.59(T)\\
		MVP(\textbf{ours})  & 190 & 83.53$\pm$0.34 & 1.21(T) & 60 & 89.25$\pm$0.23 & 0.12(T) \\
		\bottomrule
	\end{tabular} }
	\label{tab:results_diff_sizes}
\vspace{-1em}
\end{center}
\end{table}

\subsection{Results of MVP on sub-tasks of different sizes}

To evaluate the performance of MVP on sub-tasks of different sizes, we build a dataset of pruned models for $10$-classification and $3$-classification sub-tasks from CIFAR-100, of which the pruning ratio is set to $85\%$ and $95\%$ respectively.
The results are shown in Tab.\ref{tab:results_diff_sizes}. From the results we can find that When changing the size of the sub-tasks, MVP can consistently achieve comparable or better performance than SoTA methods by spending much less computation. MVP is applicable to a variety of tasks of different sizes.\looseness-1

\begin{figure*}[h]
\vspace{-1em}
     \centering
     \subfigure[Effect of Pruning Iterations]{
         \centering
         \includegraphics[width=0.4\textwidth]{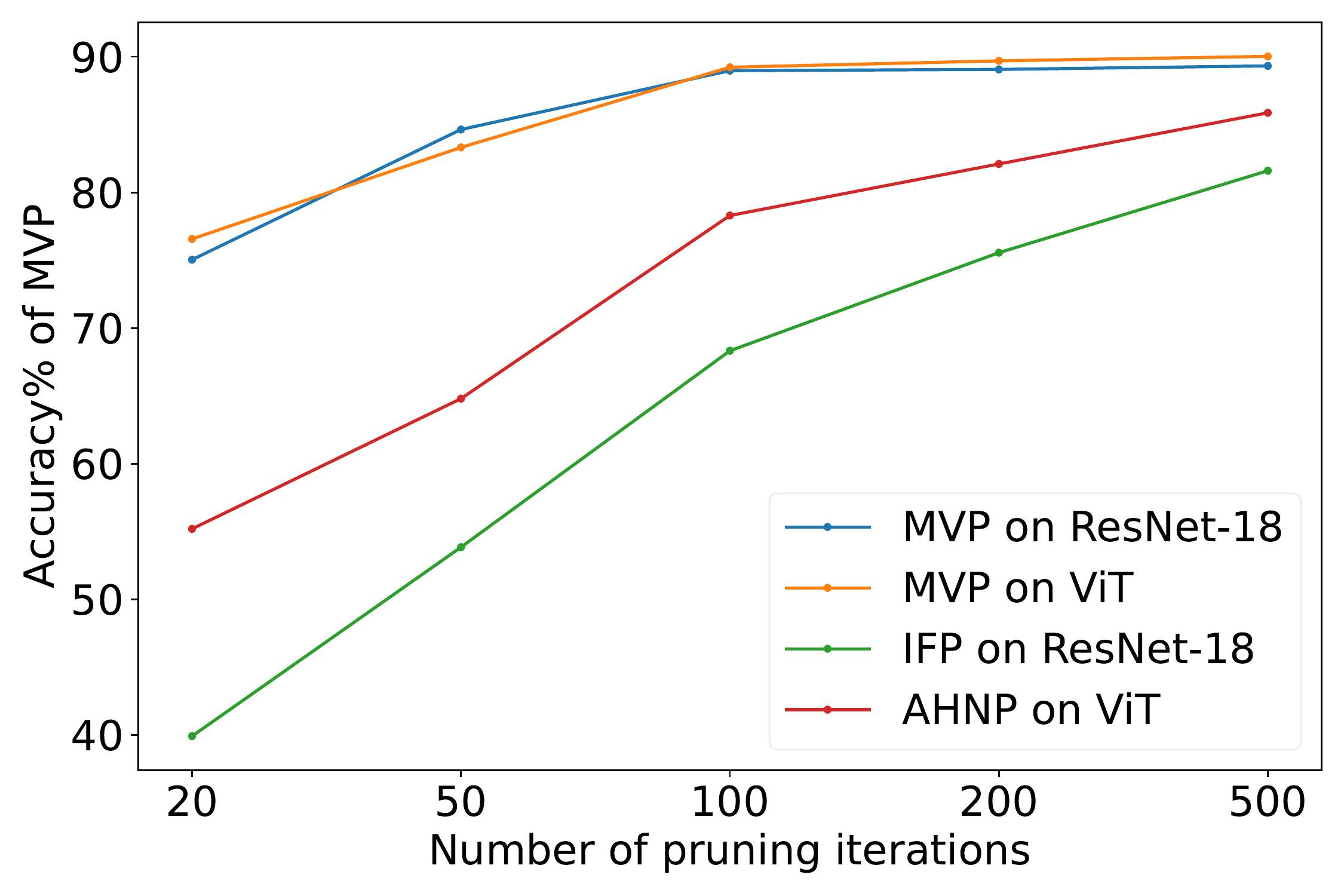}
     }
     \subfigure[Effect of Similarities between Tasks]{
         \centering
         \includegraphics[width=0.4\textwidth]{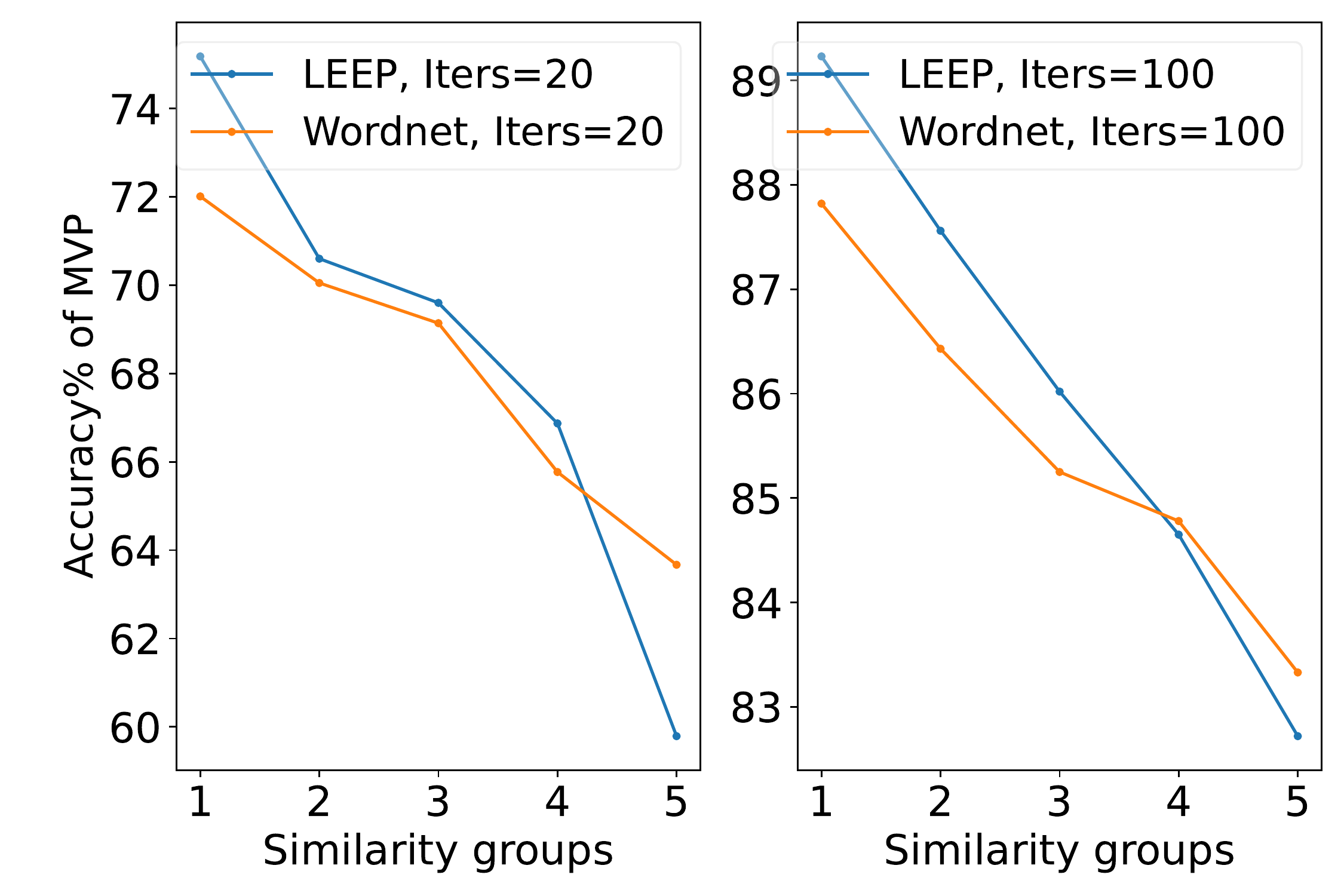}
     }
\vspace{-0.5em}
\caption{\textbf{(a)} Comparison between MVP and conventional pruning methods with different pruning iterations on different architectures. For both ResNet-18 and ViT, MVP converges much faster in a small number of iterations than conventional pruning methods.
\textbf{(b)} Comparison between LEEP score and Wordnet similarity for MVP with different pruning iterations. From similarity groups 1 to 5, the similarities between tasks decrease. For both similarity metrics, more similar tasks get better performance. LEEP score has a better ability to measure similarities between tasks than Wordnet similarity.
}
\vspace{-0.5em}
\label{fig:diff_iter_simi}
\end{figure*}

\subsection{Ablation Study}

\textbf{Effect of Iteration Numbers} 
Given a new target task and a pre-trained model, MVP can build a well-performed small model in a few iterations, demonstrating its capability in reducing adaptation cost. 
In plot (a) of Fig.~\ref{fig:diff_iter_simi}, we compare MVP with conventional pruning methods using different numbers of iterations. 
On different architectures of pre-trained models, MVP converges to a high accuracy after nearly $100$ iteration. On the contrary, the conventional pruning methods need much more iterations($>$ 500) to be comparable to MVP.
With only $\leq$ 50 pruning iterations, MVP can reach a reasonable accuracy, while conventional pruning methods perform poorly.
These imply that the initialized sub-network obtained by majority voting already contains helpful knowledge from its similar tasks to speed up the training of the pruned model.\looseness-1

\textbf{Effect of Similarities between Tasks}
MVP consistently achieves better performance when applied to nearest tasks with the highest similarities. 
In plot (b) of Fig.~\ref{fig:diff_iter_simi}, we compare the LEEP score with the Wordnet similarity and study the effect of applying MVP to neighbour tasks with different similarities. From similarity group $1$ to group $5$, the similarities between tasks decrease. 
We find that for both the two similarity metrics, the accuracy of MVP improves significantly when the similarities between tasks increase. When the pruning iterations are small($=20$), where the initialization of the sub-network is more important, the accuracy of tasks from similarity group $1$ leads to similarity group $5$ by $15\%$. Despite the accuracy of similarity group $5$ improving when the pruning iterations increase to $100$, there is still a gap of $7\%$. This result indicates that neighbour tasks with high similarities share more knowledge with the target task.
In this plot, we also find that tasks in different similarity groups classified by LEEP score show larger differences than Wordnet similarity, implying that LEEP score can better evaluate similarities between tasks. This result is consistent with our observation in the empirical study. The performance of Wordnet similarity is also good and can still be an alternative when the time and computational resources are limited.\looseness-1

\begin{figure*}[h]
     \centering
     \subfigure[IoU of Layers for IFP(Taylor Pruning)]{
         \centering
         \includegraphics[width=0.4\textwidth]{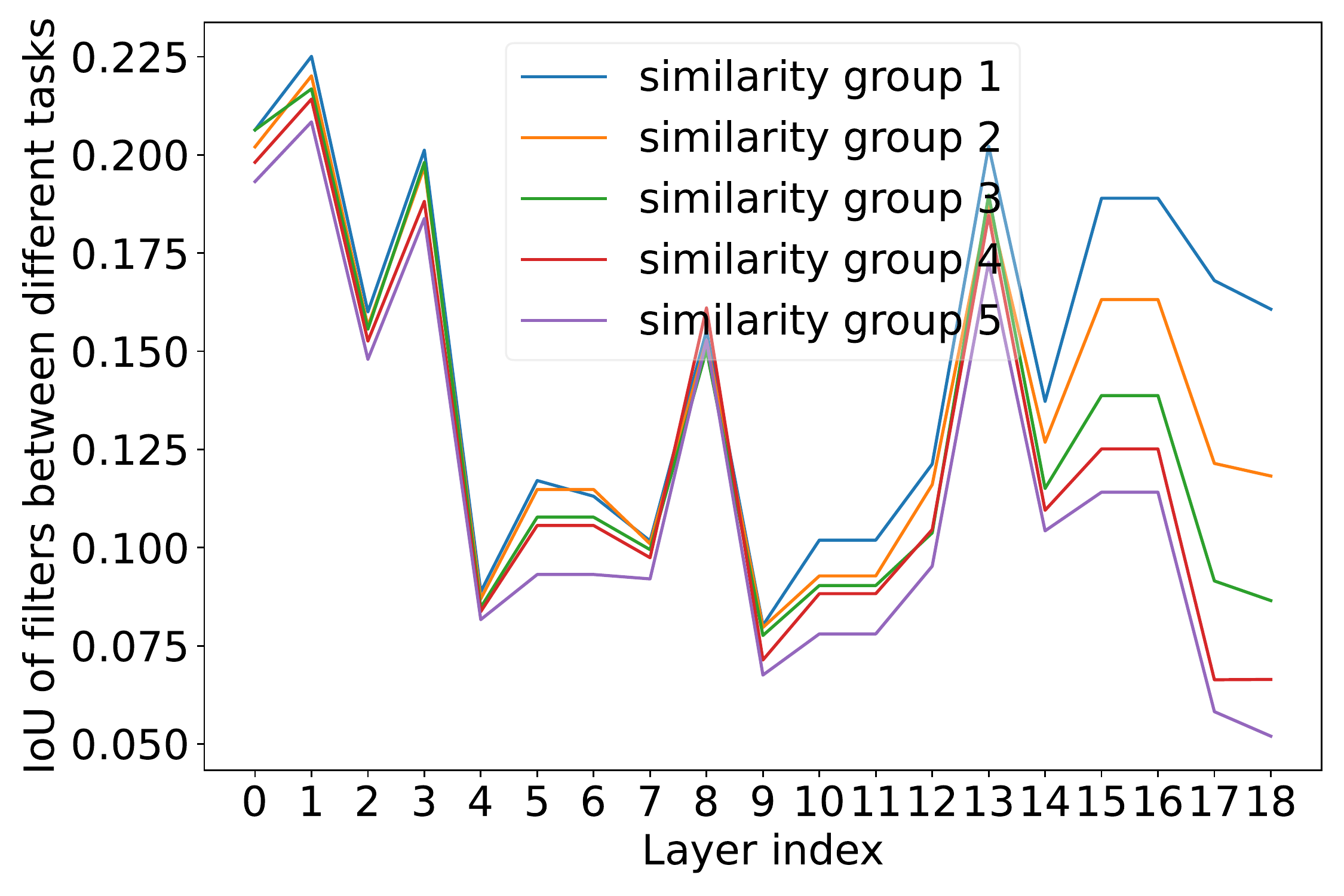}
     }
     \subfigure[Effect of Neighbour Numbers]{
         \centering
         \includegraphics[width=0.4\textwidth]{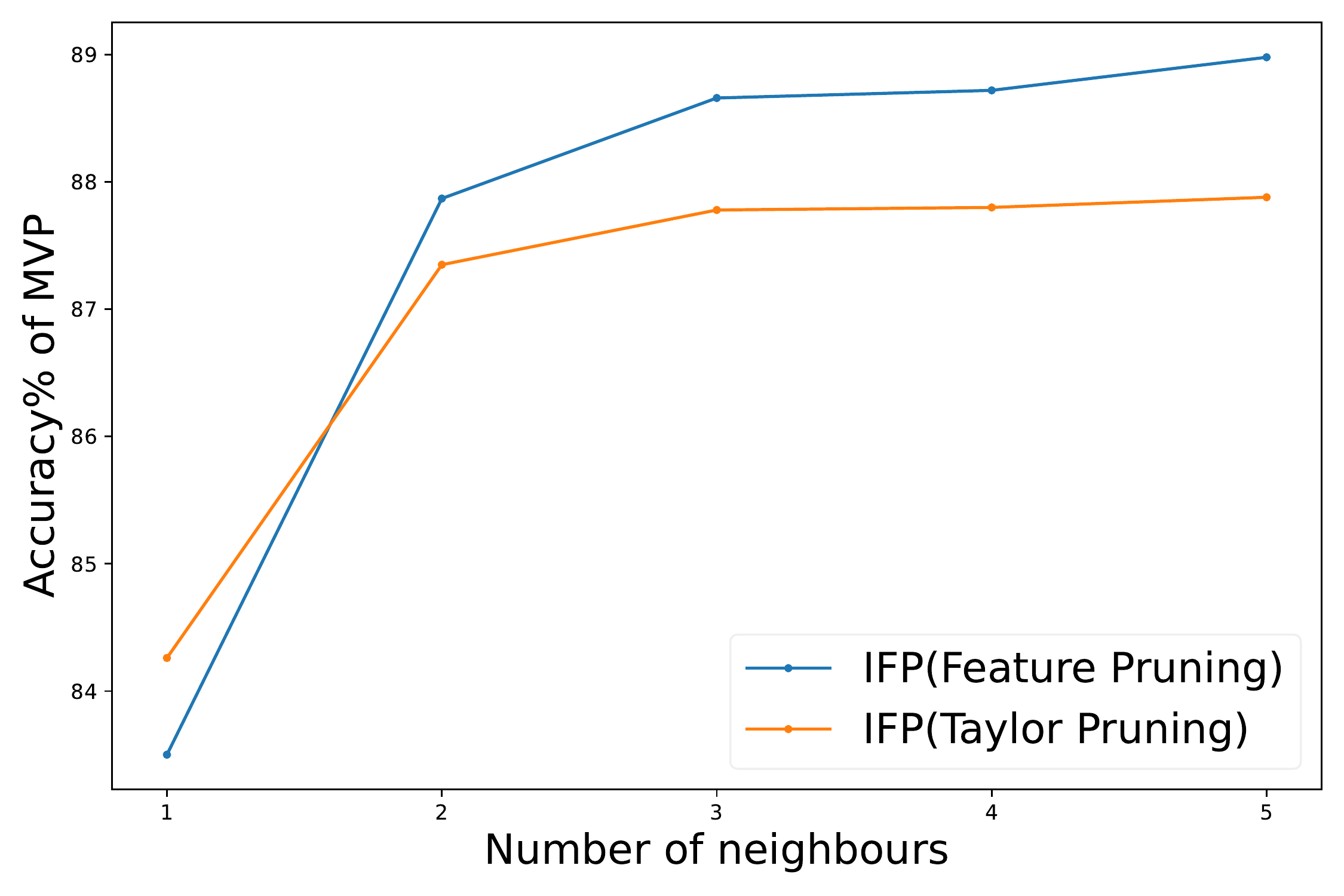}
     }
\vspace{-0.5em}
\caption{
\textbf{(a)} IoU of layers in ResNet-18 between tasks whose pruned models are extracted by IFP (Taylor Pruning) and more similar tasks also share more filters, especially in deeper layers.
\textbf{(b)} Results of applying MVP to pruned models from Activation Pruning and Taylor Pruning over different number of neighbours. MVP(neighbour number $\geq2$) can improve the performance of transfer learning(neighbour number $=1$) by a large margin when applied to pruned models extracted by different pruning methods.}
\label{fig:diff_neighbour_num}
\end{figure*}

\textbf{Comparison between Pruned Models Extracted by Different Pruning Method}
In this part, we apply MVP to pruned models extracted by Taylor Pruning~\cite{molchanov2019importance} on ResNet-18 for CIFAR-100 tasks, to prove that MVP works well on pruned models extracted by various pruning methods. Taylor Pruning measures the importance of each filter by the effect of removing this filter on final loss.
In plot (a) of Figure~\ref{fig:diff_neighbour_num}, we show the IoU of each layer for pairs of tasks with different task similarities, of which the pruned models are extracted by Taylor Pruning. Consistent with our observation in the empirical study, pruned models with higher similarities share more filters.\looseness-1

\textbf{Effect of Number of Neighbours}
In plot (b) of Figure~\ref{fig:diff_neighbour_num}, we investigate the effect of the number of neighbours for MVP. When the number $= 1$, MVP reduces to transfer learning which learns from the pruned model of a single selected similar task. In the plot, when the number of neighbours increase from $1$ to $2$, the performance improves sharply. This result implies the effectiveness of meta knowledge from different neighbours. When the number of neighbours $\geq3$, for both Activation Pruning and Taylor Pruning, the accuracy improves little, which indicates that $3$ neighbours are enough for MVP to produce a high-quality initialization.\looseness-1

\section{Conclusion}

In this paper, we study ``non-parametric meta-pruning'' problem that aims to reduce the memory and computational costs of single-task pruning, via reusing a pre-trained model and similar tasks' pruned models to find an initialization sub-network for a new task. We conduct an empirical study to investigate the relationship between task similarity and the pruned models of two tasks for different datasets and deep neural networks. The empirical study motivates a simple yet strong baseline for meta-pruning, called ``meta-vote pruning (MVP)'' (Alg.~\ref{alg:meta-pruning}). By extensive experiments on multiple tasks drawn from several datasets under different training settings, we demonstrate the advantages of MVP over other SOTA pruning methods in the region of limited computation and show its potential to reduce the carbon footprint of pruning/fine-tuning large networks for billions of edge devices and tasks.\looseness-1

\bibliographystyle{named}
\bibliography{ijcai23}

\clearpage

\section{Appendix}

\begin{algorithm}
    \SetKwInOut{Input}{Input}
    \SetKwInOut{Output}{Output}
    \SetKwInOut{Init}{Initialize}
    \Input{Pre-trained network $F(\cdot; \theta)$, Task $T$ and training set $D_T$, Hyperparameters $J,h,r,p$}
    \Init{$\Omega_\ell\leftarrow[n_\ell]$, the set of filters preserved in layer-$\ell$}
    \For{$j \gets 1$ \textbf{to} $J$}{
        \If{$j\%h = 0$ and $|\Omega_\ell| > (1-r)n_\ell$}{
            \For{$\ell \gets 1$ \textbf{to} $L-1$}{
                Prune $p\%$ of filters in $\Omega_\ell$ with the smallest importance score over $D_T$;
            }
        }
        Apply one SGD step on a mini-batch of $D_T$ to fine-tune the remained filters $\{\theta_{\ell,i}:\ell\in[L-1], i\in\Omega_\ell\}$ and $\theta_L$;
        }
\caption{\sc{Iterative Filter Pruning (IFP)}}
\label{alg:iterative_pruning}
\end{algorithm}

\subsection{Iterative Filter Pruning}
The detailed procedure of IFP is described in Algorithm~\ref{alg:iterative_pruning}.
Given a pre-trained network $F(\cdot;\theta)$ of $L$ layers (layer-$L$ is fully-connected) with parameter $\theta=\{\theta_{\ell}\}_{\ell=1:L}$ and a training set $D_T$ of a target task $T$, let $\theta_\ell=\{\theta_{\ell,i}\}_{i=1:n_{\ell}}$ denote all parameters in layer-$\ell$ composed of $\theta_{\ell,i}$ for every filter-$i$. IFP fine-tunes the model for total $J$ iterations. It prunes $p\%$ of the filters remained in each layer every $h$ iterations according to their activation values $f_{\ell,i}(x)$. It stops to prune layer-$\ell$ if reaching the targeted pruning ratio $r$.\looseness-1

\subsection{Automatic Head\&Node Pruning}
The detailed procedure of AHNP is described in Algorithm~\ref{alg:automatic_pruning}.
Given a pre-trained network $F(\cdot;\theta)$ of $L$ layers (layer-$L$ is fully-connected) with parameter $\theta=\{\theta_{\ell}\}_{\ell=1:L}$ and a training set $D_T$ of a target task $T$, let $\theta_\ell=\{\theta_{\ell,i}\}_{i=1:n_{\ell}}$ denote all parameters in layer-$\ell$ composed of $\theta_{\ell,i}$ for every head/node-$i$. $S_{\ell,i}$ denote the score for each prunable head/node-$i$ in layer-$\ell$.
AHNP fine-tunes the model and scores for total $J$ iterations. It prunes the heads/nodes if their scores are smaller than the threshold $\tau$. It stops to prune layer-$\ell$ if reaching the targeted pruning ratio $r$. Then, AHNP fine-tunes the pruned model for K iterations.\looseness-1

\begin{algorithm}
    \SetKwInOut{Input}{Input}
    \SetKwInOut{Output}{Output}
    \SetKwInOut{Init}{Initialize}
    \Input{Pre-trained network $F(\cdot; \theta)$, Task $T$ and training set $D_T$, Hyperparameters $J,K,r,\tau $}
    \Init{$\Omega_\ell\leftarrow[n_\ell]$, the set of heads/nodes preserved in layer-$\ell$. $S_{\ell,i}\leftarrow1$, the score for each prunable head/node in layer-$\ell$.}
    \For{$j \gets 1$ \textbf{to} $J$}{
        \For{$\ell \gets 1$ \textbf{to} $L-1$}{
            \For{$i\in\Omega_\ell$}{
                Prune the head/node if its score $S_{\ell,i}<\tau$;}}
        Stop pruning if reaching the target pruning ratio $r$.
        
        Apply one optimization step on a mini-batch of $D_T$ to fine-tune the remained heads/nodes and scores $\{\theta_{\ell,i},S_{\ell,i}:\ell\in[L-1], i\in\Omega_\ell\}$ and $\theta_L$;
        }
    Remove $S$, fine-tune the pruned model for K iterations on Di.
\caption{\sc{Automatic Head\&Node Pruning (AHNP)}}
\label{alg:automatic_pruning}
\end{algorithm}

\subsection{Results of MVP on Oxford Flowers-102 and Oxford-IIIT Pets}

\begin{table}[ht]

\begin{center}
\setlength{\tabcolsep}{2.pt}
    \caption{Accuracy of MVP on fine-grained tasks.}

\resizebox{\columnwidth}{!}{
	\begin{tabular}{lcccccc}
		\toprule
		\multirow{2}*{\textbf{Methods}}   & \multicolumn{3}{c}{\textbf{Oxford Flowers-102}} & \multicolumn{3}{c}{\textbf{Oxford-IIIT Pets}}  \\
		 & Iters & Acc & FLOPs & Iters & Acc & FLOPs  \\
		\midrule
		IFP & 800 & 95.20$\pm$1.39& 47.98(T)& 1000 & 77.38$\pm$0.96 & 59.33(T) \\
		IFP & 60 & 54.20$\pm$6.61 & 3.04(T) & 100 & 55.76$\pm$3.86 & 4.57(T)\\
		MVP(\textbf{ours}) & 60 & \textbf{95.40$\pm$1.34} & 0.95(T) & 100 & \textbf{78.29$\pm$0.77} & 1.58(T)\\
		\bottomrule
	\end{tabular} }
\label{tab:finegrain_results}

\end{center}
\end{table}

Similar to previous observations, MVP outperforms IFP with much more training iterations when applied to target tasks sampled from fine-grained datasets Oxford Flowers-102 and Oxford-IIIT Pets. The results indicate that MVP can build high-quality initialization, which contains fine-grained pattern information for each class in hard-to-distinguish target tasks.

\subsection{Discussion about the cost of creating a pruned model-zoo by training on hundreds of tasks}

In the experimental setting, there exists a one-time cost for preparing the model-zoo. However, like training any meta-learning model, this can facilitate many future tasks by significantly reducing their computation and required samples. Moreover, we can keep adding MVP pruned models of new tasks into this model-zoo and keep improving it in a life-long learning manner with no extra cost.

Meta-pruning moves required computation from new-task adaptation to pre-training (preparing the model-zoo). In practice, this is an even more important advantage over single-task pruning because the meta-pruning cost is offline on the server side so it is tolerable. Practitioners care more about the deployment cost of new tasks, e.g., on edge devices with limited data and computation. Because of the model-zoo, our method makes the practical deployment of model pruning more affordable.

\subsection{The difference of IoU between different similarity groups and similarity metrics}

\begin{figure}[t]
\centering
\includegraphics[width=\columnwidth]{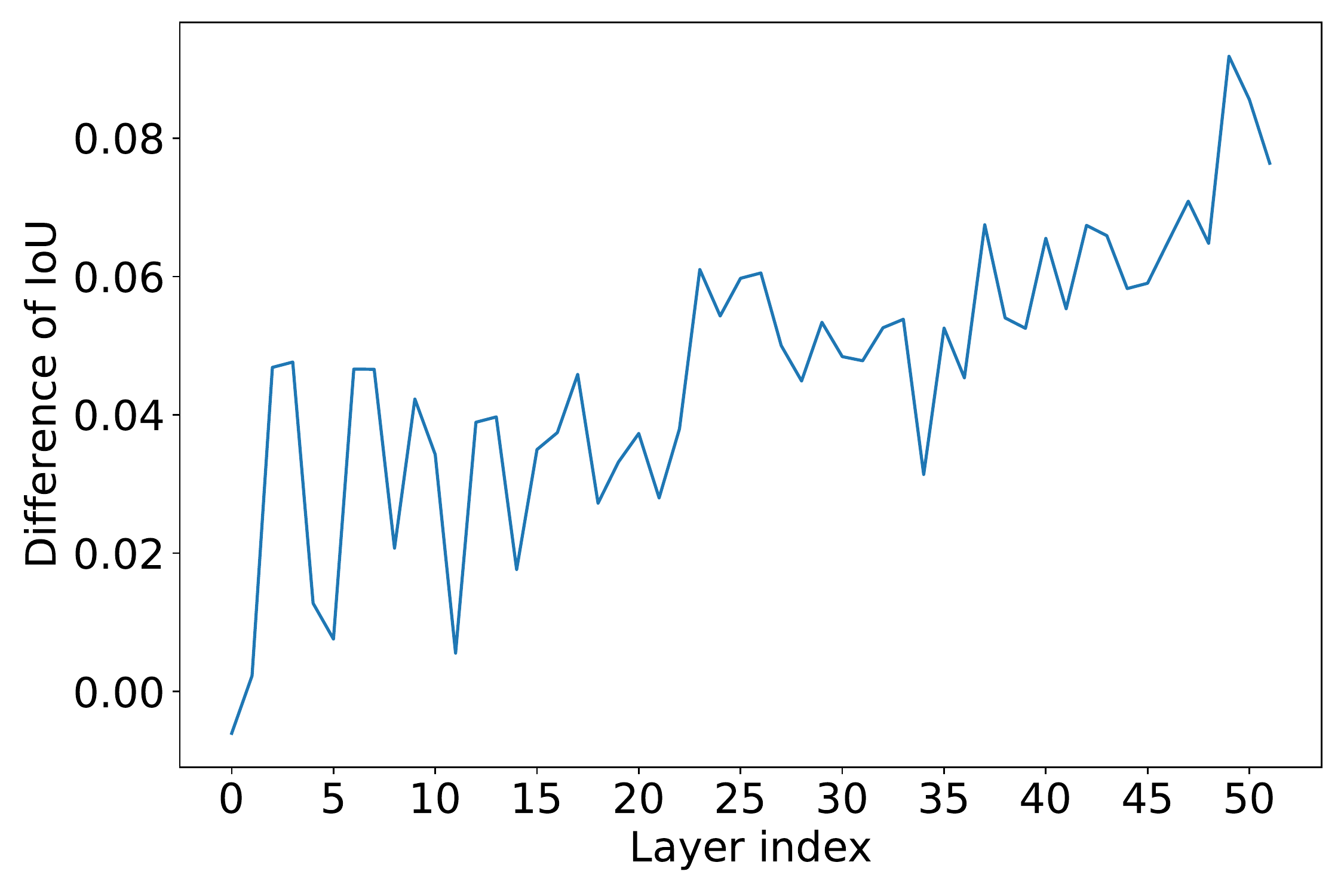}
\caption{\footnotesize The difference of IoU between tasks of similarity group 1 and similarity group 5 for ResNet-50.}
\label{fig:iou_difference}
\end{figure}

In Fig.\ref{fig:iou_difference}, we draw the difference of IoU between tasks of similarity group 1 and similarity group 5 for ResNet-50. As the layer gets deeper, the difference increases.

In Fig.2 of the paper, the average difference of IoU between similarity group 1 and similarity group 5 over all layers is $0.195$ and $0.150$ respectively for LEEP and Wordnet similarity, which has a large gap. In Fig.4 of the paper, the LEEP score performs a little better than Wordnet similarity in MVP which indicates that models with the larger IoU share more relevant parameters and LEEP has a good ability to find the nearest neighbours.

\subsection{Comparison of IoU between MVP and random pruning}

\begin{figure}[t]
\centering
\includegraphics[width=\columnwidth]{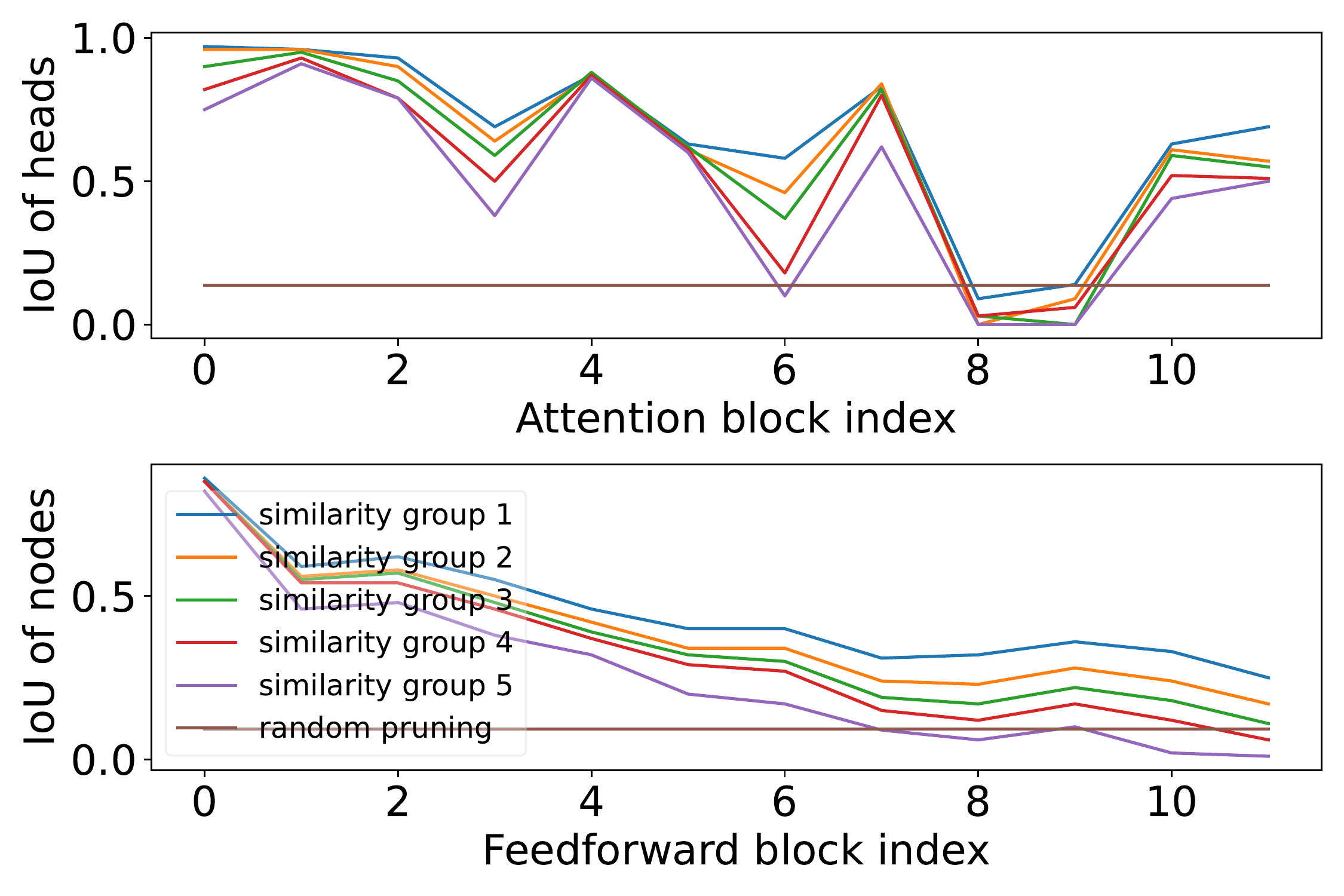}
\caption{\footnotesize Comparison of IoU between MVP and random pruning.}
\label{fig:iou_of_random}
\end{figure}

In Fig.\ref{fig:iou_of_random}, besides MVP, we also show the IoU of pruned models extracted by random pruning. When the similarity between tasks is large, the IoU of MVP is much larger than random pruning, implying that these tasks contain lots of relevant information. When the similarity between tasks is small, in the last few blocks, the IoU of nodes is similar to that of random pruning, which indicates that tasks of low similarity share little high-level information with the target task.

\end{document}